\newcommand{\CLASSINPUTtoptextmargin}{0.8in}
\documentclass[conference]{IEEEtran}
\IEEEoverridecommandlockouts
\usepackage{cite}
\usepackage{amsmath,amssymb,amsfonts}
\usepackage{algorithmic}
\usepackage{graphicx}
\usepackage{textcomp}
\usepackage{xcolor}
\usepackage{url}
\usepackage{lipsum}
\newcommand\blfootnote[1]{%
  \begingroup
  \renewcommand\thefootnote{}\footnote{#1}%
  \addtocounter{footnote}{-1}%
  \endgroup
}

\IEEEaftertitletext{\vspace{-2.0\baselineskip}}

\def\BibTeX{{\rm B\kern-.05em{\sc i\kern-.025em b}\kern-.08em
    T\kern-.1667em\lower.7ex\hbox{E}\kern-.125emX}}
\begin{document}


\title{Towards Robotic Haptic Proxies in Virtual Reality 

}

\author{\IEEEauthorblockN{Eric Godden}
\IEEEauthorblockA{\textit{Department of Electrical and Computer Engineering} \\
\textit{Queen's University}\\
Kingston, Canada \\
18eg16@queensu.ca}
\and
\IEEEauthorblockN{Matthew Pan}
\IEEEauthorblockA{\textit{Department of Electrical and Computer Engineering} \\
\textit{Queen's University}\\
Kingston, Canada \\
matthew.pan@queensu.ca}
}

\maketitle
\renewcommand{\baselinestretch}{0.88}
\begin{abstract}
This work represents the initial development of a haptic display system for increased presence in virtual experiences. The developed system creates a two-way connection between a virtual space, mediated through a virtual reality headset, and a physical space, mediated through a robotic manipulator, creating the foundation for future haptic display development using the haptic proxy framework. Here, we assesses hand-tracking performance of the Meta Quest Pro headset, examining hand tracking latency and static positional error to characterize performance of our system. 
\end{abstract}


\begin{figure}[ht]
    \centering
    \includegraphics[width=0.7\columnwidth]{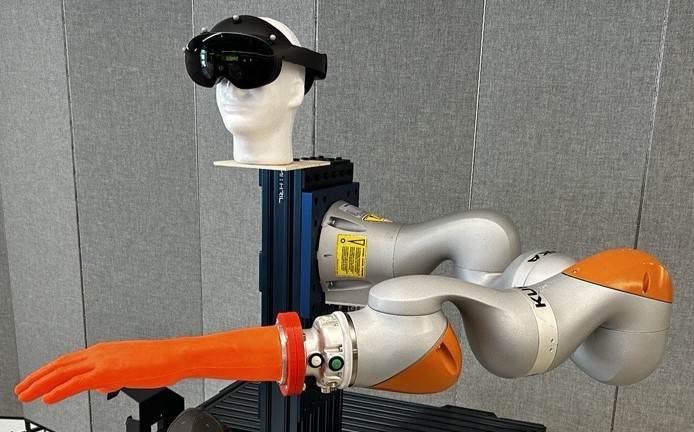}
    \caption{System setup to measure hand tracking latency of the Meta Quest Pro headset. The robot moves the replica hand in a sinusoid motion while hand pose data, as seen by the headset, is recorded. Phase differences between the robot- and headset-reported motion are used to determine latency. }
    \label{fig:test_setup}
\end{figure}

\section{Introduction}
Virtual reality (VR) development has demonstrated tremendous growth in recent years; it has changed the way users are able to interact with the digital world and has opened up a plethora of new experiences. Prior development of this technology has primarily focused on the audio-visual experience, with current products delivering near-realistic renderings of reality for these modalities. However, technology acting on the haptic modality, relating to the sense of touch and proprioception, is further behind in its development lifespan, limiting immersion in VR experiences. 


This work aims to create a haptic display system that facilitates greater presence experienced by the user within VR experiences. The system will render dynamic, physical interactions using haptic proxies coordinated through attachments to a robotic manipulator. We report on preliminary system development along with testing of hand-tracking performance that is crucial for enabling our proposed system.


\section{Background}
Presence is a foundational term within the virtual reality domain that is often tied to user experience and effectiveness of virtual simulation. Witmer et al. \cite{witmer1998measuring} describe presence as the subjective experience of being in one environment when physically in another. Haptic displays represent a broad range of devices used to provide physical stimuli to a user and play an important role in facilitating presence in virtual experiences. They enable feedback on another sensory channel and provide a pathway for immediate physical feedback \cite{dangxiao2019haptic}. In effect, they allow for a more natural mode of physical interaction, which increases the involvement and immersion experienced by the user. Additionally, they provide important enhancements to virtual reality to deliver high-presence experiences. One promising framework of haptic displays is through the use of haptic proxies: physical props that replicate virtual objects during interaction with virtual environments \cite{nilsson2021propping}, providing an untethered approach to haptic rendering of virtual objects. Robotic manipulators have proven to be successful for the co-location of physical props with their virtual counterparts in haptic proxy systems \cite{araujo2016snake}. 



\section{System}
Toward the development of a robotic haptic proxy system that aims to provide dynamic interactions with users, we take steps to integrate a robotic manipulator and VR system. \blfootnote{This work was supported by the Natural Sciences and Engineering Research Council of Canada.} The proof-of-concept integration is demonstrated through a teleoperation task where a robotic manipulator follows the user's hand trajectory as captured by an inside-out tracking system of a VR headset. 

We use a horizontally-mounted KUKA LBR iiwa 7 R800 robot manipulator that is torque-controlled in real-time at 1kHz loop rates using KUKA's proprietary Fast Robot Interface (FRI) \cite{schreiber2010fast}. To drive robot behaviours, we use a machine (Intel Core i5-2400 3.10GHz, 16GB RAM) installed with PREEMPT-RT patched Ubuntu 22.04 and ROS 2 (Humble); the lbr\_fri\_ros2\_stack was used to provide ROS 2 packages for FRI communication with the robot \cite{huber2023lbrstack}. 

The virtual environment was rendered via the Meta Quest Pro head-mounted display\footnote{https://www.meta.com/ca/quest/quest-pro/} that has a wireless connection with  Unity 3D game engine (2022.3.17f1) on a Windows 11 x64 system (AMD Ryzen 7 6800H 3.2GHz, 16GB RAM, NVidia Geforce RTX 3050 Ti) via D-Link's DWA-F18 VR Air Bridge. A feature of this headset is its ability to perform markerless hand tracking, the poses of which are accessible through Unity's XR Interaction Toolkit.\footnote{\url{https://docs.unity3d.com/Packages/com.unity.xr.interaction.toolkit@3.0/manual/index.html}} 

Communication between the VR and robot machines is performed using high-speed UDP over ethernet; data is serialized using Google's Protocol Buffer framework.\footnote{https://github.com/protocolbuffers/protobuf}



\begin{figure}[tbh!]
    \centering
    \includegraphics[width=0.9\columnwidth]{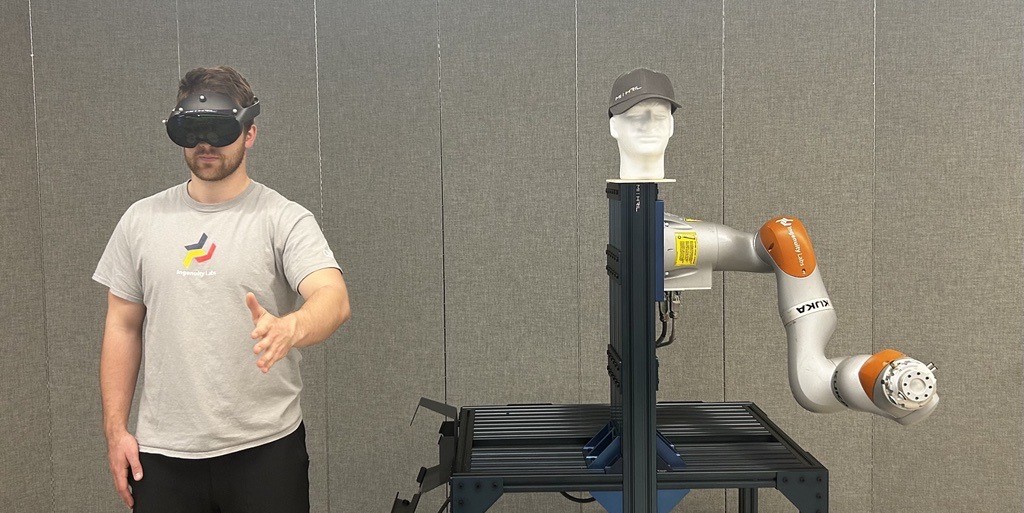}
    \caption{Teleoperation task illustrating aligned user-robot motion.}
    \label{fig:teleop}
\end{figure}




\section{Results and Discussion}
\subsection{Latency}
As a measure of system performance, we tested the latency and positional error in the headset's hand-tracking. To quantify latency, we devised an experimental setup where a replica of a human hand was attached as the end effector of the robot, and the headset was affixed near the base of the robot as shown in Figure \ref{fig:test_setup}. The phase difference between the headset's reported motion of the hand and the manipulator's sinusoidal motion of the first joint was found to determine the hand-tracking latency of the system. 
Analysis of the hand-tracking latency, as seen by the robot, showed a bimodal distribution of phase differences between robot-reported and headset-reported end-effector poses: phase differences seen when the sinusoid motion first starts from rest were much greater (m: 177.4 ms, sd: 78.0 ms) than when sinusoid motion is underway (m: 27.8 ms, sd: 13.7 ms) over 10 trials of recorded motion. We noticed in some cases that the headset's reported pose led the robot-reported pose; we suspect that these results are due to the headset predicting future poses of the hands based on current states to compensate for latencies in tracking - this is confirmed by Meta's literature \cite{noauthor_powered_nodate}. From prior work, any visual and haptic latency above 16 ms will likely be discernable by users and adversely affect immersion~\cite{lee_discrimination_2009}; our current observations show latencies above this threshold. Thus, in creating robotic haptic proxies, we will need to use kinematic model that predicts a user’s motion into the future to mitigate such latencies. 

\subsection{Position Accuracy}
A motion capture system consisting of 12 Vicon Vero cameras (Tracker 3.9.0) was used to provide ground-truth reference to determine hand-tracking positional error. The average hand-tracking positional error and standard deviation of the headset was determined to be 1.30 cm and 1.48 cm, respectively. This accuracy result was determined through the comparison of 30 tracked static positions of the replica hand's index finger tip. For haptic encounters, this error may result in minor but perceptible mismatches between visual and haptic feedback, but we plan on conducting more studies to determine system efficacy as a result of such errors. 

\subsection{Teleoperation Task}
As a first demonstration of our system setup, we conceived a teleoperation task, as seen in Figure \ref{fig:teleop}, where the robot's end-effector follows the movement of a user's hand as tracked by the headset. The hand's pose is tracked with respect to the headset's frame and sent over UDP. Because the robot's configuration and mounting are similar to that of a human arm, the conversion of the hand pose from within the headset's space to the robot's space is trivial. A 2x scalar multiplication on Cartesian coordinates is applied to account for the approximate ratio of arm lengths of the robot to the human. An analytic inverse kinematics solver is used to convert Cartesian poses into joint angles; motion is planned using the Ruckig motion generator \cite{berscheid2021jerk} and sent to the robot via FRI. The target pose of the robot end-effector is updated in the motion generator as soon as new hand pose data is received at a rate of 90Hz. In parallel, the robot's current state and joint angles are read via a ROS topic provided by lbr\_fri\_ros2\_stack, and are relayed over UDP to the VR system. The VR system then updates its virtual rendering of the manipulator to provide the user with visual feedback of the manipulator's motion in response to the user's hand motion.



\section{Conclusions and Future Work}
In this work, the initial development of a robotic haptic proxy system for dynamic interactions with users in VR was presented. The system established a connection between VR and robot manipulator subsystems, which proved successful in a teleoperation task. Hand-tracking latency analysis demonstrated delays above discernible thresholds and static positional error was determined. Latency reduction and the understanding of the impact of current levels of positional error are essential for future integration into the desired framework.



\bibliographystyle{IEEEtran}
\bibliography{main}

\end{document}